\documentclass{article}

 \usepackage[preprint]{neurips_2019}

\usepackage[utf8]{inputenc} % allow utf-8 input
\usepackage[T1]{fontenc}    % use 8-bit T1 fonts
\usepackage{hyperref}       % hyperlinks
\usepackage{url}            % simple URL typesetting
\usepackage{booktabs}       % professional-quality tables
\usepackage{amsfonts}       % blackboard math symbols
\usepackage{nicefrac}       % compact symbols for 1/2, etc.
\usepackage{microtype}      % microtypography
\usepackage{graphicx}
\usepackage{subfigure}
\usepackage{multirow}
\usepackage{makecell}
\usepackage{epsfig,amsmath,amssymb,amsthm,array}
\usepackage{epstopdf}

\newtheorem{theorem}{Theorem}

\newtheorem{algorithm}{Algorithm}
\newtheorem{remark}{Remark}

\usepackage{url}
\usepackage{algorithmicx,algorithm}
\title{Training Artificial Neural Networks by Generalized Likelihood Ratio Method: Exploring Brain-like Learning to Improve Robustness}

\author{%
Li Xiao \\
 Institute of Computing Technology\\
  Chinese Academy of Science\\
  \texttt{xiaoli@ict.ac.cn} \\
   \And
Yijie Peng \\
 Peking University \\
   \texttt{pengyijie@pku.edu.cn} \\
   \AND
Jeff Hong\\
Fudan University \\
   \texttt{hong\_liu@fudan.edu.cn} \\
   \And
Zewu Ke\\
 Institute of Computing Technology\\
  Chinese Academy of Science\\
  \texttt{18795206856@163.com} \\
   \And
Shuhuai Yang\\
Peking University\\
  \texttt{aaronshyang@pku.edu.cn} \\
}

\begin{document}

\maketitle

% this must go after the closing bracket ] following \twocolumn[ ...

% This command actually creates the footnote in the first column
% listing the affiliations and the copyright notice.
% The command takes one argument, which is text to display at the start of the footnote.
% The \icmlEqualContribution command is standard text for equal contribution.
% Remove it (just {}) if you do not need this facility.

%\printAffiliationsAndNotice{}  % leave blank if no need to mention equal contribution
%\printAffiliationsAndNotice{\icmlEqualContribution} % otherwise use the standard text.

\begin{abstract}
 In this work, we propose a generalized likelihood ratio method capable of training the artificial neural networks with some biological brain-like mechanisms,.e.g., (a) learning by the loss value, (b) learning via neurons with discontinuous activation and loss functions. The traditional back propagation method cannot train the artificial neural networks with aforementioned brain-like learning mechanisms. Numerical results show that the robustness of various artificial neural networks trained by the new method is significantly improved when the input data is affected by both the natural noises and adversarial attacks.   Code is available at: \url{https://github.com/LX-doctorAI/GLR_ADV} .
\end{abstract}

\section{Introduction}
\label{introduction}	 
Artificial neural network (ANN) has been used as a universal classifier. In recent years, there have been tremendous successes in applying ANNs to image processing, speech recognition, game, and medical diagnosis (\citealp{Kaiming2016Deep,Graves2013Speech,Silver2016Mastering,Gulshan2016Development,Esteva2017dermatologist}) . In ANN, the inputs such as texts and images are turned into a vector, and each neuron performs a nonlinear transformation on the input vector. A deep learning ANN typically contains multiple layers of convoluted neurons. This complicated machinery maps the input space to the target space. There are synaptic weights in each neuron to be adapted to the surrounding environment based on the loss between the ANN output and target data. The back propagation (BP) method has been the most widely used technique to train ANNs. However, the BP method requires the loss function and activation function to be smooth in ANNs, which limits the capability of ANNs to fit well with the surrounding environments. 

Recent work in deep learning has demonstrated that ANNs can be more easily confused by small noises added to the images via snowing, blurring, and pixelation than human being (\citealp{Dietterich2018benchmarking,Shankar2018classifier,Weiss2018transform,Dan2018Using}). Moreover, ANNs are vulnerable to adversarial attacks, where a very small perturbation of the inputs can drastically alter the classification result (\citealp{Szegedy2014Intriguing,Goodfellow2014Explaining,Carlini2017Towards,Seyed2016Deepfool,Bastani2016Measuring}). In contrast, the adversarial phenomenon rarely happens for human being (\citealp{Gamaleldin2018adversarialhuman}). 

Then some interesting questions arise: what are the differences between biological neural networks in the human brains and ANNs? Can we borrow some mechanisms from the biological brain neural networks to improve the robustness of ANNs? There are some noticeable differences between the neurons in human brain and the neurons used in traditional ANNs (\citealp{theoreticalneuroscience}). First, the activation of the brain neuron is via an electric impulse, which can be captured by a threshold activation function, and there exists a (neuronal) noise in each neuron, whereas the traditional ANN uses continuous activation functions, e.g., Sigmoid and ReLu, and a deterministic nonlinear transformation. Second, human brain perceives an object as a specific category, e.g., dog or cat, which means that the loss function capturing the mechanism of a human brain should be a discontinuous zero-one function, whereas the loss functions in ANNs are smooth, e,g., the cross-entropy function. Furthermore,  
the brain neuron network is effected directly by the electronic signal sent by a sensory system and the chemical signal from the endocrine, therefore the biological brain functions like learning from the loss value itself rather than the gradient of the loss, which the BP method uses.
	
In this work, a generalized likelihood ratio (GLR) method is proposed to train ANNs with neuronal noises. Unlike the BP method, GLR trains ANNs directly by the loss value, rather than the gradient of the loss.  GLR does not differentiate the sample path of the loss, and it can train ANNs with discontinuous activation and loss functions. Therefore, the new training method generalizes the scope of ANNs to be used in practice, which allows some brain-like mechanisms, i.e., (a) learning by the loss value and (b) learning via neurons with discontinuous activation functions and neuronal noises. The complexity in calculating the GLR estimator is also simpler than the BP method, because there is no need to calculate the backward propagation for the derivatives of the error signals. 

The GLR method is a recent advance in stochastic gradient estimation studied actively in the area of simulation optimization (\citealp{asmussen2007stochastic}, \citealp{fu2015gradient}, \citealp{peng2015discontinuity}). Infinitesimal perturbation analysis (IPA) and the likelihood ratio (LR) method are two classic unbiased stochastic gradient estimation techniques (\citealp{ho1991discrete}, \citealp{glasserman1991gradient}, \citealp{hong2009estimating}, \citealp{rubinstein1993discrete}, \citealp{pflug1996}, \citealp{heidergott2010weak}). IPA allows the parameters in the performance function but requires the continuity (differentiability) of the performance function; LR does not allow the parameters in the performance function, whereas it does not require continuity of the performance function. The GLR method extends two classic methods to a setting allowing both the parameters in the performance function and discontinuous performance function, so that it can be applied to train the parameters in discontinuous ANNs. 

We test the classification results of various trained ANNs when the input data is corrupted by both the natural noises and the adversarial attacks. The robustness of all ANNs trained by the GLR method is significantly improved compared with the ANN with the Sigmoid activation function and cross-entropy loss function trained by the BP method.

\section{Method}
\subsection{Setup and Background}
\begin{figure}[!h]
	\begin{center}
		\includegraphics[scale=0.3]{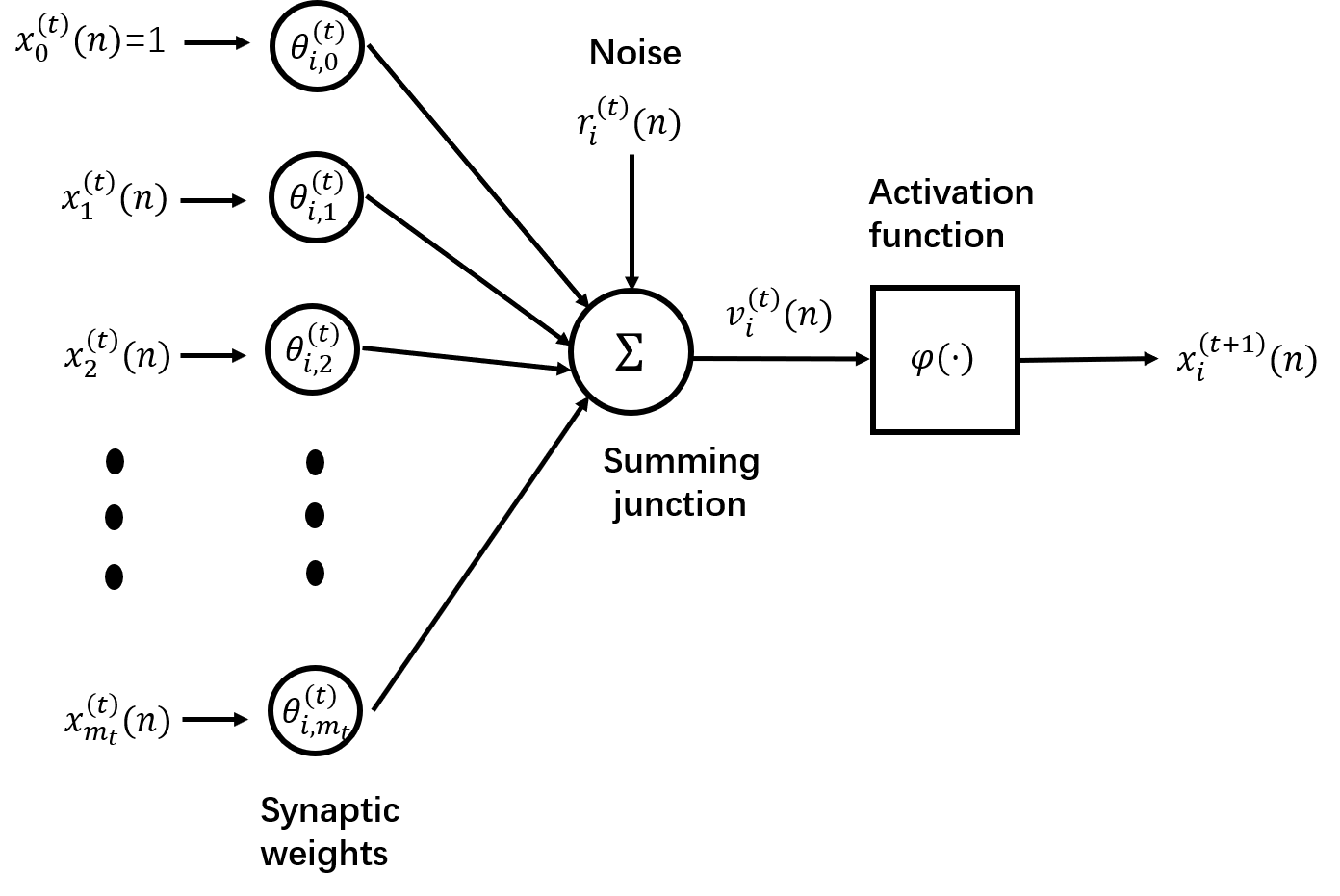}
		\caption{Structure of a neuron.}
		\label{pic3} 
\end{center} \end{figure}

Suppose we have inputs $(x_1^{(1)}(n),\ldots,x_{m_1}^{(1)}(n))$, $n=1,\ldots, N$. 
For the $n$th input, the $i$th output of the $t$th level of neurons is given by 
\begin{align*}
x_i^{(t+1)}(n):=\varphi\left(v_i^{(t)}(n)\right),\quad v_i^{(t)}(n):=\sum_{j=0}^{m_t} \theta_{i,j}^{(t)}x_j^{(t)}(n)+r^{(t)}_i(n), \quad i=1,\ldots, m_{t+1},
\end{align*}
where $x_j^{(t)}(n)$ is the $j$th input of the $t$th level of neurons ($j$th output of the $(t-1)$th level of neurons), $\theta_{i,j}^{(t)}$ is a synaptic weight, $r^{(t)}_i(n)$ is a noise, $v_i^{(t)}(n)$ is the $i$th signal, and $\varphi$ is the activation function. The synaptic weights $\theta_{i,j}^{(t)}$, $j=0,\ldots,m_t$, are the parameters to be trained in the ANN. It is required that $x_{0}^{(t)}(n)\equiv 1$, and $\theta_{i,0}^{(t)}$ is called bias. Figure \ref{pic3} illustrates the structure of a neuron in the ANN. 

\begin{figure}[!h]
	\begin{center}
		\includegraphics[scale=0.5]{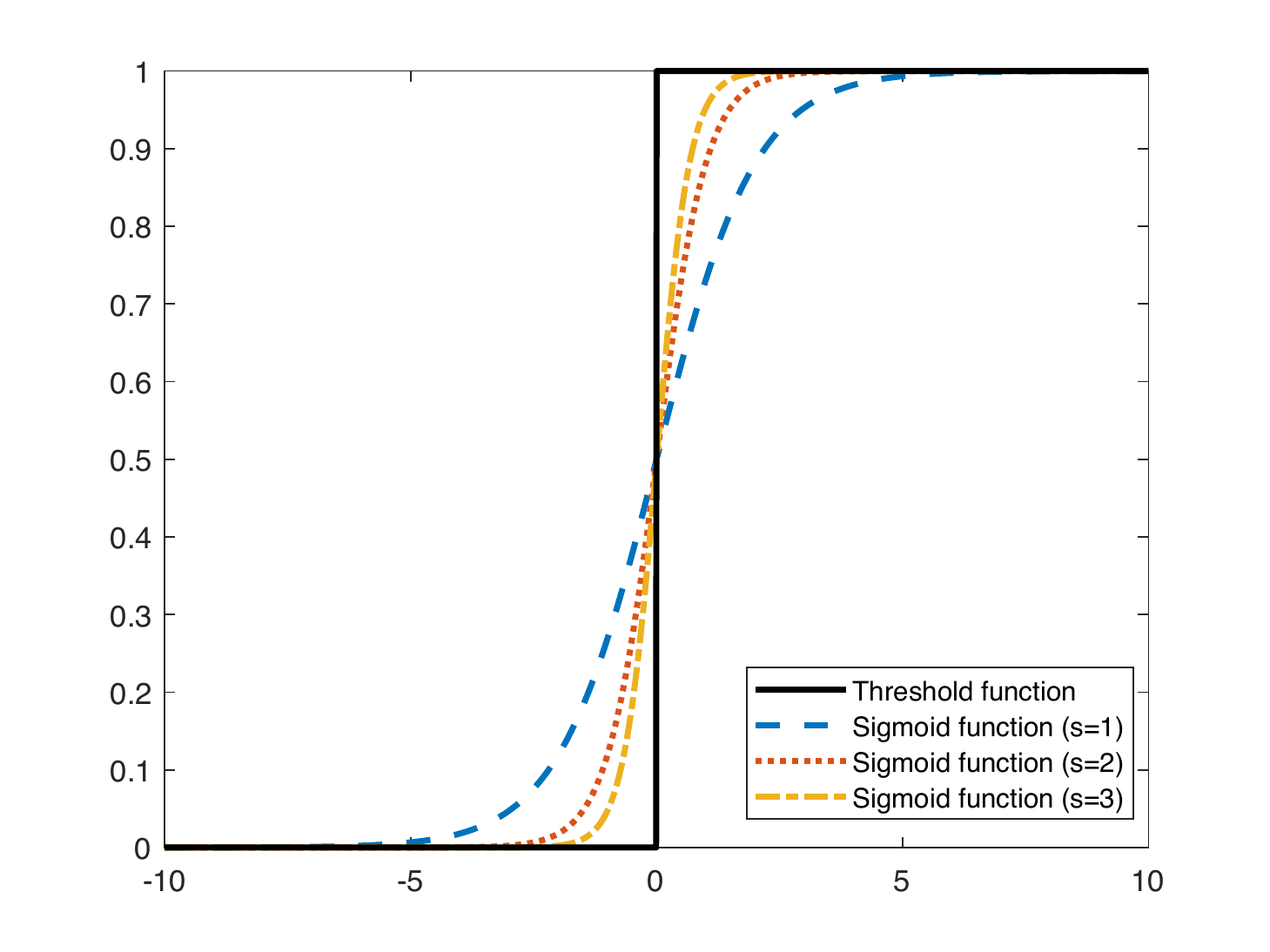}
		\caption{ Threshold activation function and Sigmoid activation functions.}
		\label{pic1} 
\end{center} \end{figure}

The classic ANN does not include the noise, i.e., $r_i^{(t)}(n)\equiv 0$, and the ANN considered in our work generalizes the classic one by adding a (random) noise to the neurons. The activation function $\varphi$ is a nonlinear function. The Sigmoid function %and ReLu 
is a popular activation function defined by 
\begin{align*}
\varphi_s(v):=1/(1+\exp(-sv)),
\end{align*}
where $s>0$ is a constant. %, and the ReLu is defined by 
%\begin{align*}
%\varphi_r(v):=\max(v,0)~.
%\end{align*}
The Sigmoid function is smooth. Notice that with parameter $s$ increasing to infinity, the Sigmoid function converges to a threshold function, i.e., 
\begin{align*}
\lim_{s\to\infty}\varphi_s(v)=\varphi_o(v):=
\begin{cases}
1\quad  &\mbox{if $v> 0$},
\\
0\quad  &\mbox{if $v<0$},
\end{cases}\quad a.e.
\end{align*}
In Figure \ref{pic1}, we can see the curves of the Sigmoid functions with different parameters and the threshold function. 

Suppose the ANN has $\tau$ layers of neurons, and $\vec{X}^{(\tau)}(n):=(x^{(\tau)}_1(n),\ldots,x^{(\tau)}_{m_\tau}(n))$ is the output vector of the ANN given the $n$th input data. Let $\vec{O}(n):=(o_1(n),\ldots,o_{m_\tau}(n))$ be the real observation vector given the $n$th input data, and $L(\vec{X}^{(\tau)}(n),\vec{O}(n))$ be a loss function of the outputs of ANN and observations.  In  classification, a popular loss function is the cross-entropy loss function given by 
\begin{align*}
L_c(\vec{X}^{(\tau)}(n),\vec{O}(n)):=-\sum_{i=1}^{m_\tau} o_i(n)\log \left(p_i(\vec{X}^{(\tau)}(n))\right),
\end{align*}
where 
$$p_i(\vec{X}^{(\tau)}(n)):=\frac{\exp\left(x_i^{(\tau)}(n)\right)}{\sum_{j=1}^{m_\tau}\exp\left(x_j^{(\tau)}(n)\right)},\quad i=1,\ldots, m_\tau~.$$
 The functions $p_i$, $i=1,\ldots,m_\tau$, are called softmax functions.  Note that the cross-entropy loss function is smooth. Alternatively, we can also use the following 0-1 loss function: 
 \begin{align*}
L_o(\vec{X}^{(\tau)}(n),\vec{O}(n)):={\bf1}\left\{ \arg\max_{i=1,\ldots,m_\tau} p_i(\vec{X}^{(\tau)}(n))=  \arg\max_{i=1,\ldots,m_\tau} o_i(\vec{X}^{(\tau)}(n))\right\}~.
 \end{align*} 

To train the ANN, we want to minimize the expected loss: 
\begin{align*}
\mathcal{E}(\theta)=\mathbb{E}\left[L(\vec{X}^{(\tau)}(n),\vec{O}(n)) \right],
\end{align*}
where $\theta$ is a vector containing all synaptic weights. To solve the optimization,  the stochastic approximation (SA) (\citealp{yin2003stochastic}) is applied  
\begin{equation}
\theta(n)= \theta(n-1)-\lambda_n G(n)  ,
\label{eq0}
\end{equation}
where $\lambda_n$ is the learning rate and $ G ( n ) $ is an unbiased stochastic gradient estimator of $ \mathcal{E}(\theta) $, i.e., 
\begin{equation}
\mathbb{E}[ G ( n ) ]=  \left . \nabla_\theta\mathcal{E}(\theta )  \right |_{ \theta= \theta(n-1)} ~.
\label{eq1}
\end{equation}
In training ANN, the BP algorithm is the most popular stochastic gradient estimator (\citealp{haykin2009neural}): 
	\begin{align*}
B_{a,b}^{(l)}(n):=\delta_{a}^{(l+1)}(n) x_{b}^{(l)}(n),
\end{align*}
where $B_{a,b}^{(l)}(n)$ is an unbiased stochastic derivative estimator with respect to synaptic weight $\theta_{a,b}^{(l)}$, and 
\begin{equation*}
\begin{aligned}
\delta_i^{(t)}(n):=\begin{cases}
e_{i}^{(\tau)}(n)\varphi'\left(v_i^{(\tau-1)}(n)\right),\quad  &\mbox{if $t=\tau$},
\\
\varphi'\left(v_i^{(t-1)}(n)\right)\left( \sum_{j=1}^{m_{t}}\theta_{j,i}^{(t)} \delta_j^{(t+1)}(n)\right),\quad  &\mbox{if $t<\tau$},
\end{cases}
\end{aligned}
\end{equation*}
with the error signal $e_{i}^{(\tau)}(n)$ defined by 
\begin{align*}
e_{i}^{(\tau)}(n):=\frac{\partial L(\vec{X}^{\tau}(n),\vec{O}(n))}{\partial x_{i}^{(\tau)}(n)}~.
\end{align*}

\subsection{Generalized Likelihood Ratio method}
 \cite{peng2019stochastic} prove that the BP algorithm is mathematically equivalent to IPA, but the computational complexity of BP is lower. BP directly differentiates the sample path of the output, so it requires the sample path of the output is Lipchitz continuous and differentiable almost surely. Therefore, BP cannot deal with the stochastic gradient estimation for ANN with discontinuous activation function and loss function. ANN used in our work may contain a discontinuous activation or loss function, which leads to a discontinuous sample path of the output. A push-out LR method proposed in \cite{peng2019stochastic} can deal with the stochastic gradient estimation for ANN with discontinuous activation function and loss function. Here we derive a GLR estimator which coincides with the push-out LR estimator for the first-order gradient in \cite{peng2019stochastic} under a special case with the Sigmoid activation and a bounded loss function with a bounded gradient. The derivation shows the connection between the GLR method and  the BP method. 
We construct an ANN with the Sigmoid activation function:

 \begin{align*}
  y_i^{(t+1)}(n):=\varphi_s\left(u_i^{(t)}(n)\right),\quad u_i^{(t)}(n):=\sum_{j=0}^{m_t} \theta_{i,j}^{(t)}y_j^{(t)}(n)+r^{(t)}_i(n), \quad i=1,\ldots, m_{t+1},
  \end{align*}
  and an ANN with the threshold activation function: 
  \begin{align*}
  z_i^{(t+1)}(n):=\varphi_o\left(\eta_i^{(t)}(n)\right),\quad \eta_i^{(t)}(n):=\sum_{j=0}^{m_t} \theta_{i,j}^{(t)}z_j^{(t)}(n)+r^{(t)}_i(n), \quad i=1,\ldots, m_{t+1}~.
  \end{align*}	

\begin{theorem} \label{th2} 
	Assuming that density function $f_{a,l}(\cdot)$ of the noise $r_a^{(l)}(n)$ (added to the $a$-th output of the $(l-1)$-th level of neurons)  is differentiable and $\lim_{r\to\pm \infty}f_{a,l}(r)=0$, and  loss function $L(\cdot)$ is bounded and with a bounded gradient w.r.t  $\vec{X}^{(\tau)}(n)$,
	we have 
	\begin{align*}
	\frac{\partial }{\partial \theta_{a,b}^{(l)}}\mathbb{E}\left[ L(\vec{Z}(n),\vec{O}(n))\right]=	\mathbb{E}\left[- L(\vec{Z}(n),\vec{O}(n))~z_b^{(l)}(n)\frac{\partial \log f_{a,l}(r_a^{(l)}(n))}{\partial r_a^{(l)}(n)}\right]~.
	\end{align*}
\end{theorem}
\proof 
We have 
\begin{align*}
\frac{\partial L(\vec{Y}^{(\tau)}(n),\vec{O}(n))}{\partial \theta_{a,b}^{(l)}}=\sum_{i=1}^{m_\tau} \frac{\partial L(\vec{Y}^{(\tau)}(n),\vec{O}(n))}{\partial y_i^{(\tau)}(n)} \frac{\partial y_i^{(\tau)}(n)}{\partial \theta_{a,b}^{(l)}},
\end{align*}
where  $\vec{Y}^{(\tau)}(n):=(y_1^{(\tau)}(n),\ldots,y_{m_\tau}^{(\tau)}(n))$, and
\begin{align*}
\frac{\partial y_i^{(t+1)}(n)}{\partial \theta_{a,b}^{(l)}}=\varphi'_s\left(u_i^{(t)}(n)\right)\frac{\partial u_i^{(t)}(n)}{\partial \theta_{a,b}^{(l)}},\quad \frac{\partial u_i^{(t)}(n)}{\partial \theta_{a,b}^{(l)}}=\sum_{j=0}^{m_t} \left(\frac{\partial \theta_{i,j}^{(t)}}{\partial \theta_{a,b}^{(l)}} y_j^{(t)}(n)+\theta_{i,j}^{(t)}\frac{\partial y_j^{(t)}(n)}{\partial \theta_{a,b}^{(l)}} \right)~.
\end{align*}
In addition, we have 
\begin{align*}
\frac{\partial L(\vec{Y}^{(\tau)}(n),\vec{O}(n))}{\partial \theta_{a,b}^{(l)}}=\sum_{i=1}^{m_\tau}\frac{\partial L(\vec{Y}^{(\tau)}(n),\vec{O}(n))}{\partial y^{(\tau)}_i} \frac{\partial y_i^{(\tau)}(n)}{\partial \theta_{a,b}^{(l)}},
\end{align*}
where 
\begin{align*}
\frac{\partial y_i^{(t+1)}(n)}{\partial r_{a}^{(l)}}=\varphi'\left(u_i^{(t)}(n)\right)\frac{\partial u_i^{(t)}(n)}{\partial r_{a}^{(l)}},\qquad \frac{\partial u_i^{(t)}(n)}{\partial r_{a}^{(l)}}=\sum_{j=0}^{m_t} \left(\theta_{i,j}^{(t)}\frac{\partial y_j^{(t)}(n)}{\partial r_{a}^{(l)}}+\frac{\partial r_i^{(t)}(n)}{\partial r_{a}^{(l)}} \right) ~.
\end{align*}
Notice that 
\begin{align*}
 \frac{\partial u_a^{(l)}(n)}{\partial \theta_{a,b}^{(l)}}= y_b^{(l)}(n),\qquad  \frac{\partial u_a^{(l)}(n)}{\partial r_{a}^{(l)}}=1~.
\end{align*}
It is straightforward to show 
\begin{align*}
\left.\frac{\partial L(\vec{Y}^{(\tau)}(n),\vec{O}(n))}{\partial \theta_{a,b}^{(l)}}\right/\frac{\partial L(\vec{Y}^{(\tau)}(n),\vec{O}(n))}{\partial r_{a}^{(l)}}=y_b^{(l)}(n)~.
\end{align*}
Then,
\begin{align*}
&\frac{\partial \mathbb{E}\left[L(\vec{Y}^{(\tau)}(n),\vec{O}(n))\right]}{\partial \theta_{a,b}^{(l)}}=\mathbb{E}\left[\frac{\partial L(\vec{Y}^{(\tau)}(n),\vec{O}(n))}{\partial \theta_{a,b}^{(l)}}\right]\\
=&\mathbb{E}\left[\int_{\mathbb{R}}\frac{\partial L(\vec{Y}^{(\tau)}(n),\vec{O}(n))}{\partial \theta_{a,b}^{(l)}}f_{i,l}(r_a^{(l)})~ dr_a^{(l)}\right]\\
=&\mathbb{E}\left[\int_{\mathbb{R}}\frac{\partial L(\vec{Y}^{(\tau)}(n),\vec{O}(n))}{\partial r_{a}^{(l)}} \left(\left.\frac{\partial L(\vec{Y}^{(\tau)}(n),\vec{O}(n))}{\partial \theta_{a,b}^{(l)}}\right/\frac{\partial L(\vec{Y}^{(\tau)}(n),\vec{O}(n))}{\partial r_{a}^{(l)}}\right) f_{i,l}(r_a^{(l)})~ dr_a^{(l)}\right]\\
=&\mathbb{E}\left[\int_{\mathbb{R}}y_b^{(l)}(n) \frac{\partial L(\vec{Y}^{(\tau)}(n),\vec{O}(n))}{\partial r_{a}^{(l)}}  f_{i,l}(r_a^{(l)})~ dr_a^{(l)}\right]~.
\end{align*}
The interchange of derivative and expectation in the first equality can be justified by the dominated convergence theorem by noticing that the gradient of loss function $L(\cdot)$ w.r.t. $\vec{Y}^{(\tau)}(n)$ and the gradient of $\vec{Y}^{(\tau)}(n)$ w.r.t. $\theta_{a,b}^{(l)}$ are bounded. 
By integration by parts, 
\begin{align*}
&\mathbb{E}\left[\int_{\mathbb{R}} y_b^{(l)}(n) \frac{\partial L(\vec{Y}^{(\tau)}(n),\vec{O}(n))}{\partial r_{a}^{(l)}}  f_{i,l}(r_a^{(l)})~ dr_a^{(l)}\right]\\
=&\mathbb{E}\left[\left.y_b^{(l)}(n)~ L(\vec{Y}^{(\tau)}(n),\vec{O}(n))~ f_{i,l}(r_a^{(l)})\right|_{r_a^{(l)}=-\infty}^{\infty}-\int_{\mathbb{R}}y_b^{(l)}(n) L(\vec{Y}^{(\tau)}(n),\vec{O}(n))\frac{\partial f_{i,l}(r_a^{(l)})}{\partial r_{a}^{(l)}}  ~ dr_a^{(l)}\right]\\
=&\mathbb{E}\left[ - L(\vec{Y}^{(\tau)}(n),\vec{O}(n))~y_b^{(l)}(n)~\frac{\partial \log f_{i,l}(r_a^{(l)})}{\partial r_{a}^{(l)}}\right],
\end{align*}
where te first term is zero on the right hand side of the first equality because the loss function $L(\cdot)$ and $y_b^{(l)}(n)$ are bounded.
By taking limit, 
\begin{align*}
\lim_{s\to\infty}&\mathbb{E}\left[ - L(\vec{Y}^{(\tau)}(n),\vec{O}(n))~y_b^{(l)}(n)~\frac{\partial \log f_{i,l}(r_a^{(l)})}{\partial r_{a}^{(l)}}\right]\\
=&\mathbb{E}\left[ \lim_{s\to\infty}- L(\vec{Y}^{(\tau)}(n),\vec{O}(n))~y_b^{(l)}(n)~\frac{\partial \log f_{i,l}(r_a^{(l)})}{\partial r_{a}^{(l)}}\right]\\
=&\mathbb{E}\left[-  L(\vec{Z}^{(\tau)}(n),\vec{O}(n))~z_b^{(l)}(n)~\frac{\partial \log f_{i,l}(r_a^{(l)})}{\partial r_{a}^{(l)}}\right],
\end{align*}
where $\vec{Z}^{(\tau)}(n):=(z_1^{(\tau)}(n),\ldots,z_{m_\tau}^{(\tau)}(n))$ and the interchange of limit and expectation in the first equality can be justified by the dominated convergence theorem by noticing that loss function $L(\cdot)$ is bounded and $y_b^{(l)}(n)$ is uniformly bounded in $s$. Moreover, for $t<l$, 
$$\lim_{s\to\infty} y_i^{(t)}(n)=z_i^{(t)}(n),~~a.s.,\quad i=1,\ldots,m_t,$$ 
for a bounded neighborhood $\Theta_{a,b}^{(l)}$ containing $\theta_{a,b}^{(l)}$,
$$\lim_{s\to\infty} \sup_{ \theta_{a,b}^{(l)}\in \Theta_{a,b}^{(l)}}\left|y_b^{(l)}(n)-z_b^{(l)}(n)\right|=0,~~a.s.,$$ 
and 
$$\lim_{s\to\infty} \sup_{ \theta_{a,b}^{(l)}\in \Theta_{a,b}^{(l)}}\left|L(\vec{Y}^{(\tau)}(n),\vec{O}(n))-L(\vec{Z}^{(\tau)}(n),\vec{O}(n))\right|=0,~~a.s.,$$
which further leads to  
\begin{align}\label{uc}
\lim_{s\to\infty}\sup_{ \theta_{a,b}^{(l)}\in \Theta_{a,b}^{(l)}}\left|\mathbb{E}\left[ \left( L(\vec{Y}^{(\tau)}(n),\vec{O}(n))y_b^{(l)}(n)-L(\vec{Z}^{(\tau)}(n),\vec{O}(n))z_b^{(l)}(n)\right)\frac{\partial \log f_{i,l}(r_a^{(l)})}{\partial r_{a}^{(l)}}\right]\right|=0.
\end{align}
Summarizing the results above,  
\begin{align*}
\frac{\partial \mathbb{E}\left[L(\vec{Z}^{(\tau)}(n),\vec{O}(n))\right]}{\partial \theta_{a,b}^{(l)}}=&\frac{\partial }{\partial \theta_{a,b}^{(l)}}\lim_{s\to\infty} \mathbb{E}\left[L(\vec{Y}^{(\tau)}(n),\vec{O}(n))\right]\\
=&\lim_{s\to\infty}\frac{\partial }{\partial \theta_{a,b}^{(l)}} \mathbb{E}\left[L(\vec{Y}^{(\tau)}(n),\vec{O}(n))\right]\\
=&\lim_{s\to\infty}\mathbb{E}\left[ - L(\vec{Y}^{(\tau)}(n),\vec{O}(n))~y_b^{(l)}(n)~\frac{\partial \log f_{i,l}(r_a^{(l)})}{\partial r_{a}^{(l)}}\right],
\end{align*}
where the interchange of limit and derivative in the second equality is justified by uniform convergence (\ref{uc}). This proves the theorem. 
\endproof

\begin{remark}
	\cite{peng2019stochastic} show that for an ANN with certain smoothness in activation and loss functions,
	$$B_{a,b}^{(l)}(n)=\frac{\partial L(\vec{X}^{(\tau)}(n),\vec{O}(n))}{\partial \theta_{a,b}^{(l)}}~.$$
	The GLR estimator is defined by
	\begin{equation}
	L_{a,b}^{(l)}(n):=L(\vec{X}(n),\vec{O}(n))~\omega_{a,b}^{(l)}(n),
	\label{eq4}
	\end{equation}
	where
	\begin{equation*} 
	\omega_{a,b}^{(l)}(n):=-x_b^{(l)}(n)\frac{\partial \log f_{a,l}(r_a^{(l)}(n))}{\partial r_a^{(l)}(n)}~.
	\label{eq5}
	\end{equation*}	
	From the proof of Theorem \ref{th2}, we can see that for an ANN under certain regularity conditions, the estimator of the BP algorithm and the GLR estimator can be linked via integration by parts. For an ANN with a threshold activation function, the GLR estimator is derived by first smoothing the threshold activation function, which becomes the Sigmoid function, then integration by parts, and last taking limit to retrieve the threshold activation in the derivative estimator. These three components have also been used to derive the GLR method in a general framework (\citealp{peng2015discontinuity}), where we can find the smoothing technique is applied to a general discontinuous sample performance function without actually explicitly constructing the smoothing function. The GLR method can be generalized to deal with stochastic gradient estimation or even higher order gradient for the ANN with more general discontinuous activation and loss functions, which can be found in \cite{peng2019stochastic}. 
\end{remark}

\begin{remark} The BP method differentiates the loss and transmits the error signal from the output layer backward throughout the entire ANN via the chain rule of the derivative, whereas in Eq.\ref{eq4}, the GLR method does not differentiate the loss and directly uses the loss function scaled by a weight function, which can be viewed as an interaction between the interior mechanism of ANN and the loss in a surrounding environment, to train the ANN.  
\end{remark}
 
\begin{remark}
 The BP method is computationally efficient because it only requires simulating a forward function propagation and backward error propagation for once, and the derivatives w.r.t. all synaptic weights $\theta_{i,j}^{(t)}$, $j=1,\ldots,m_t$, $i=1,\ldots,m_{t+1}$, $t=1,\ldots,\tau-1$, are estimated. The GLR method is even faster than BP, since its computation only contains one forward function propagation for estimating the derivatives w.r.t. all parameters. For a Gaussian random noise $r_i^{(t)}$ with zero mean and variance $\sigma_{i,t}^2$, we have 
 \begin{equation*}
 \frac{\partial \log f_{i,t}(r_i^{(t)})}{\partial r_i^{(t)}}=-\frac{r_i^{(t)}}{\sigma_{i,t}^2}~. 
 \label{eq6}
 \end{equation*} 
\end{remark}
 
\subsection{Implementation Details}
In implementation, we add a Gaussian noise with zero mean and certain variance to each neuron. Then, the GLR gradient estimator used in Eq.\ref{eq4} for a synaptic weight associated with the signal from the $b$-th neuron  at the $l$-th level to the $a$-th neuron at the $(l+1)$-th level is
\begin{small}
\begin{equation}
\frac{L(\vec{X}(n),\vec{O}(n))x_b^{(l)}(n)r_a^{(l)}}{\sigma_{a,l}^2}~.
\label{eq7}
\end{equation}
\end{small}
Here $x_b^{(l)}$ is the signal from the $b$-th neuron at the $l$-th level and $r_a^{(l)}$ is the noise of the $a$-th neuron at the $(l+1)$-th level, $L(\vec{X}(n),\vec{O}(n))$ is the loss. For simplicity, we choose a common variance $\sigma^2$ for the noises in all neurons. The training procedure by the GLR method is summarized in \textbf{\emph{Algorithm \ref{alg1}}}.

\begin{algorithm}[h]
\begin{small}
\caption{Training procedure by GLR} 
{\bf Setup:} Input $\vec{X}^{(1)}(n)$, observations $\vec{O}(n)$, and variance $\sigma^2$.

{\bf Step One:} Calculate loss output $L(\vec{X}(n),\vec{O}(n))$ and the GLR gradient $G(n)$ in Eq.\ref{eq1} via Eq.\ref{eq7}.

{\bf Iterations:}  Replicate the above procedure $K$ times to generate i.i.d. gradient estimates $G_1{(n)},G_2{(n)},...,G_K{(n)}$.

{\bf Output:}  An average of GLR gradient estimates  $\frac{1}{K}\sum_{i=1}^{K}G_i{(n)}$, which is used in Eq.\ref{eq0} for updating parameter.
\label{alg1}
\end{small}
\end{algorithm}

\section{Experiments}
\subsection{Dataset Preparation and Network Structure}
\begin{figure}[h]
  	\begin{center}
  		\includegraphics[scale=0.15]{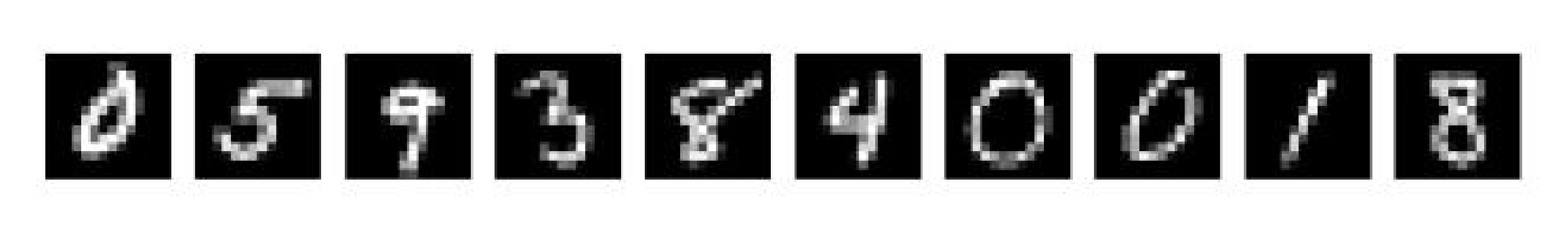}
  		\caption{ Images of the numbers in the MNIST dataset.}
  		\label{pic4} 
  \end{center} \end{figure}

We test the performance of the GLR method for training ANNs in the example of identifying the numbers from 0-9 of the MNIST dataset, where  there are  $10000$ images in total. The images are split into the training set and testing set in a 6:4 ratio. Each image is resized to be a $14\times14$-pixels vector for facilitating the training. The appearance of the images is shown in Figure.\ref{pic4}.
  
\begin{figure}[h]
%	\begin{center}
\centering
\subfigure[]{\includegraphics[scale=0.25]{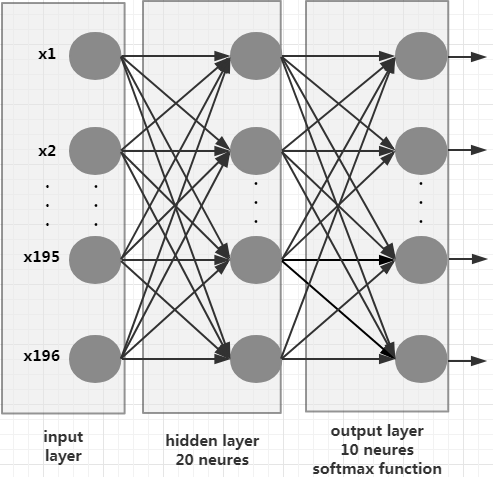}}
\subfigure[]{\includegraphics[scale=0.25]{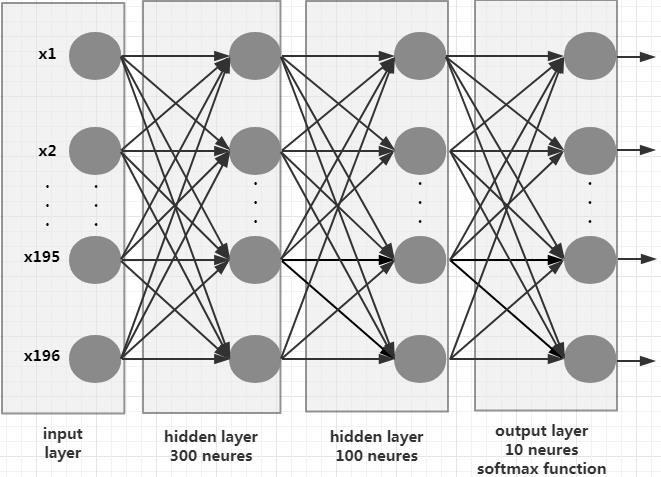}}
		\caption{(a) Structure of ANN trained by the GLR method; (b) Structure of ANN for generating adversarial samples with two hidden layers.}
		\label{pic5} 
%\end{center} 
\end{figure}

 The ANN to be trained in the experiments have three layers: an input layer, a hidden layer, and an output layer. The structure of the ANN is depicted in Figure.\ref{pic5}(a). The dimension of the input layer is $196$, the same as the size of the image. The hidden layer has $20$ neurons, and the output layer has $10$ neurons representing $10$ numbers. 
 %The bias term is an independent node set as $1$ in the input layer and hidden layer. 
  The integer value of the label needs to be converted into a $10$-unit array as the target of the ANN. For example, when the label is $2$, the target vector should be $[0,0,1,0,0,0,0,0,0,0]$. The operations between the layers are illustrated in Figure.\ref{pic3}. The inputs of the input layer and hidden layer first go through linear operations with Gaussian noises added on and then nonlinear activation functions are operated. The bias term in our ANN is set to be $1$ at the head of each input array. 
 
\subsection{Training Procedure} 
\begin{figure}[h]
 \begin{center}
\centering
\subfigure[]{\includegraphics[scale=0.11]{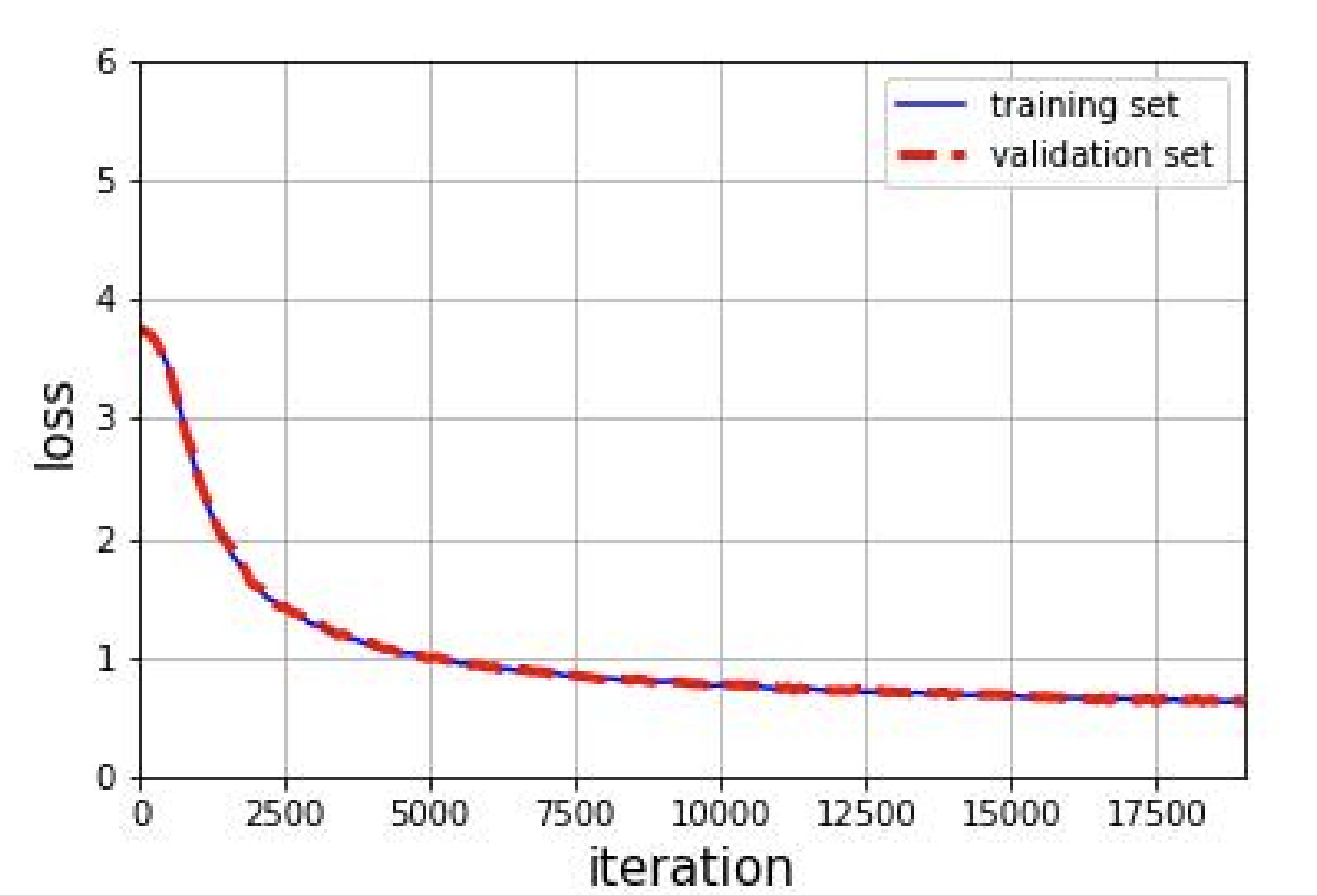}}
\subfigure[]{\includegraphics[scale=0.125]{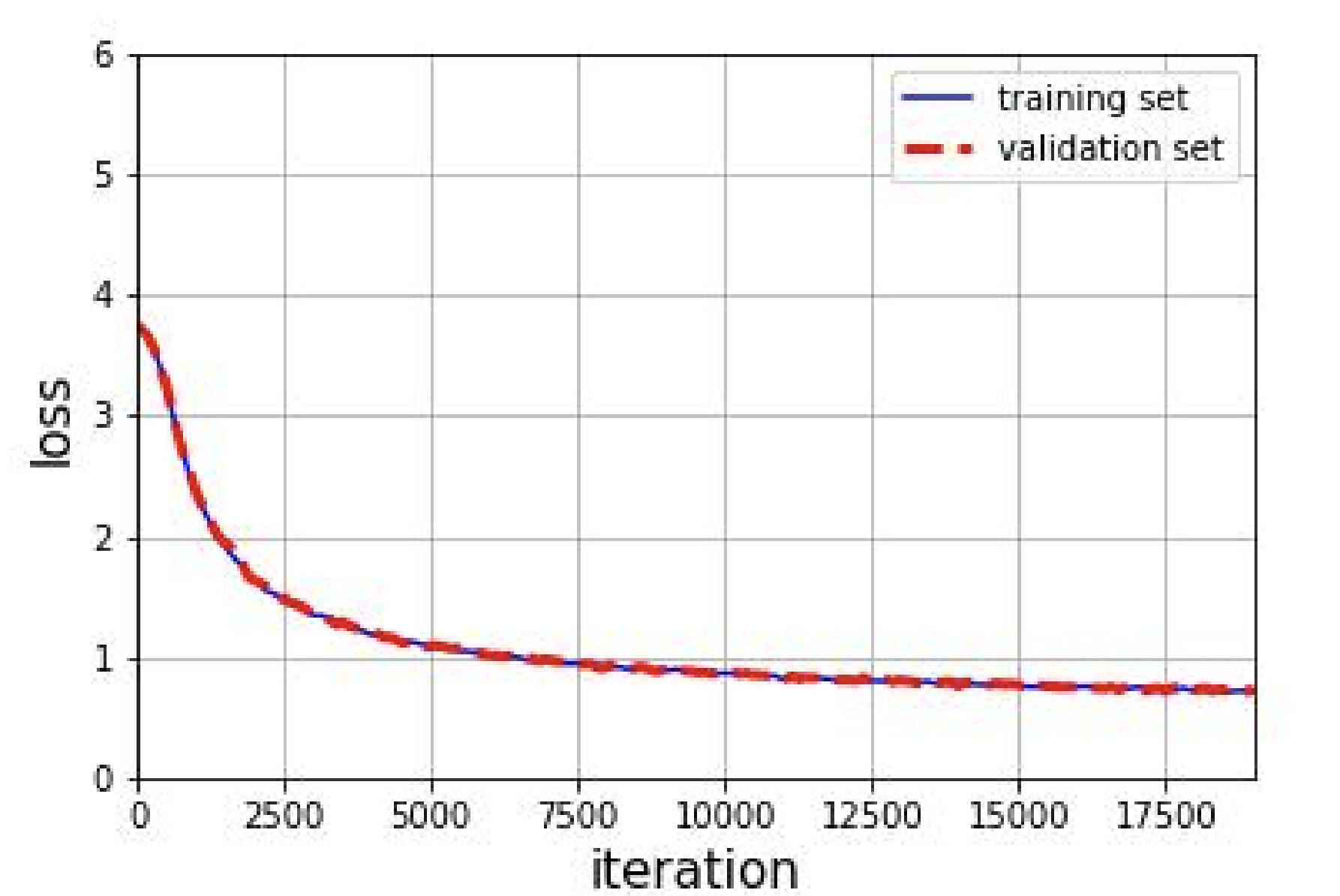}}
		\caption{Losses in the training and validation sets of a single hidden layer ANN trained by GLR with activation functions as Sigmoid in (a) and threshold in (b), respectively.}
		\label{pic6} 
\end{center} 
\end{figure}

The number of replications in \textbf{\emph{Algorithm \ref{alg1}}} is set as $K=10000$. We apply the SA in Eq.\ref{eq0} with mini-batches and the batch size is set as $25$, which takes about $12$ seconds to run in python in a desktop with Intel i7-6700 CPU @ 3.40 GHz for each iteration. Each epoch contains $1680$ iterations which takes about five hours to run. The step size is set as $0.1$ and the noise variance is set as $\sigma^2=4$. \cite{peng2019stochastic} present many training results of the GLR method on classifying the handwritten data included in the python {\it sklearn} package.  In Figure.\ref{pic6}, we show the training and validation errors of ANNs with the Sigmoid and threshold activation functions (as plotted in Figure.\ref{pic1}) in the MNIST dataset, and the errors of ANNs converge fast after $12$ epochs ($\approx20000$ iterations).

\subsection{Robustness to Adversarial Attacks}
\subsubsection{Adversarial Samples} 
\begin{figure}[h]
\begin{center}
\centering
\includegraphics[scale=0.2]{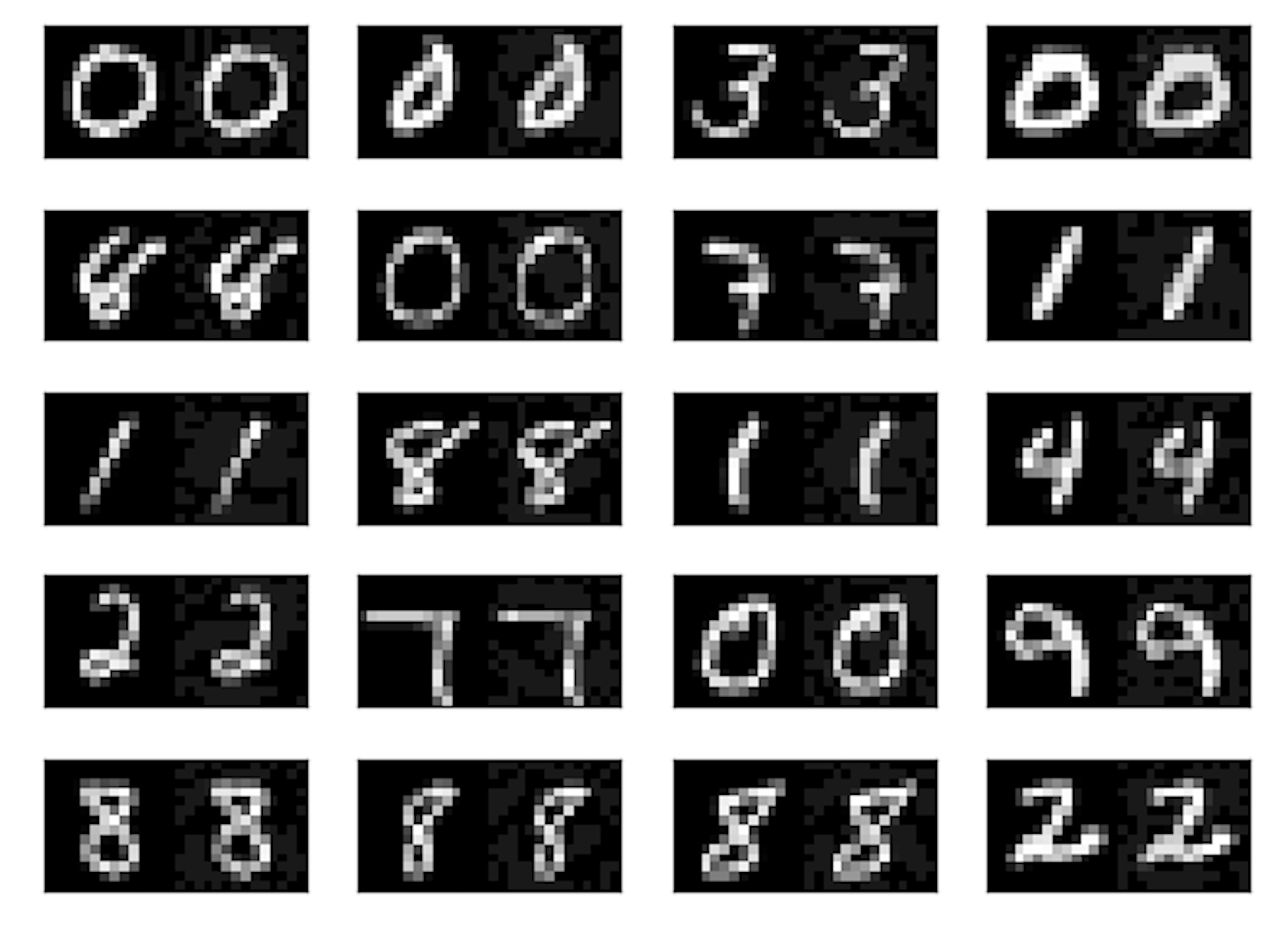}
		\caption{Samples of the paired images of the original samples (left) and  adversarial samples (right). The adversarial samples are generated by the FGSM. }
		\label{pic7} 
\end{center} 
\end{figure}

We generate the adversarial samples by an ANN with two hidden layers (as depicted in (b) of Figure.\ref{pic5}). The ANNs are trained and validated by the BP method. The limited-memory Broyden–Fletcher–Goldfarb–Shanno  (L-BFGS)(\citealp{Szegedy2014Intriguing}) and fast gradient sign method (FGSM)(\citealp{Goodfellow2014Explaining}) are used to generate the adversarial samples of $5000$ images randomly chosen from the testing set. The adversarial samples of the images generated by FGSM are shown in Figure.\ref{pic7}.

\subsubsection{Adversarial Test}
Adversarial test is performed on several single hidden layer ANNs  (Figure.\ref{pic5}(a)) with different activation and loss functions. The accuracy is measured by the percentage of correct predictions over all adversarial samples. The ANN with the same structure in (a) of Figure.\ref{pic5} trained by BP is used as the baseline for comparisons. Table.\ref{table3} presents the results when adversarial samples are generated by the ANN with two hidden layers (Figure.\ref{pic5}(b)). The accuracies of the prediction on the original samples and the adversarial samples generated by the aforementioned two methods are reported.
   \begin{table}[h]
     \begin{small}
	\centering
	\begin{tabular}{| c|  c| c| c| }\hline
	 \textbf{Activations $+$ Entropy }&  \textbf{Orig}  & \textbf{Adv\_L\_BFGS}  &\textbf{Adv\_FGSM}  \\\hline
	 Sigmoid (trained by BP)  & $0.96$ & $0.57$ & $0.28$ \\\hline
	 Sigmoid & $0.94$ &$0.77$  &$0.45$ \\\hline
	 Threshold & $0.93$ & $0.73$ &$0.52$ \\\hline
       $y=|x|$ & $0.94$ & $0.78$ & $0.53$\\\hline
		 \textbf{Activations $+$  0-1 loss } &  \textbf{Orig}  & \textbf{Adv\_L\_BFGS}  &\textbf{Adv\_FGSM}  \\\hline
	 Sigmoid &$0.84$  &$0.76$  & $0.58$ \\\hline
	Threshold & $0.83$ & $0.72$ & $0.57$\\\hline
	\end{tabular}
	\vspace*{0.1in}
	\caption{Adversarial tests for ANNs with different activation and loss functions trained by GLR. The adversarial samples are generated by an ANN with two hidden layers.  \textbf{Orig}  means the accuracy tested on original samples. \textbf{Adv\_L\_BFGS} means the accuracy tested on samples generated by the L\_BFGS method. \textbf{Adv\_FGSM} means the accuracy tested on samples generated by FGSM. }
	\label{table3}
	\end{small}
\end{table}

Besides the Sigmoid and threshold activation functions, three other discontinuous activation functions % shown in Figure.\ref{pic8} 
are also used in the experiments. Each test runs for 12 epochs and all of them demonstrate high accuracies on predicting the original samples after training. 
We also test the performance of an ANN with 0-1 loss function (the loss is 0 for a correct prediction and 1 otherwise), trained by the GLR. The ANN with 0-1 loss converges slower than the classic cross entropy loss (see Table.\ref{table3}), so we run 24 epochs in training.

In Tables.\ref{table3} , an ANN with one hidden layer trained by the BP method reaches an accuracy of $0.96$ in predicting the original samples. However,  the accuracy of the same ANN reduces dramatically to $0.57$ and $0.28$ in predicting the adversarial samples generated by L\_BFGS and FGSM, respectively. This substantiates an observation in literature, i.e., the same adversarial samples can effectively attack ANNs under different architectures. %It has been well-known that the output loss of an ANN typically has many local optima and saddle points.  A conjecture on why this adversarial phenomenon happens is that the parameters in the ANN trained by the BP method may be stuck in a local optimum of the output loss with a very steep slope around the neighborhood or a saddle point. Therefore, a slight perturbation in the input data could  lead to a significant changes in the outputs, which in turn causes the dramatic change in classification. 

All ANNs with the cross-entropy loss function trained by the GLR method achieve accuracies in predicting original samples comparable to the ANN trained by the BP method  (above $93\%$ in accuracy). Notice that GLR can train the ANNs with discontinuous activation functions, e.g., threshold function, and discontinuous loss functions, e.g, 0-1 loss, which cannot be handled by BP. Moreover, the ANNs trained by the GLR method have much higher accuracies (about $20\%$ increase) in predicting adversarial samples compared to the ANN trained by the BP method.  Another interesting observation is that although the ANNs with 0-1 loss only reach accuracies less than 90\%  in predicting original samples, they might lead to even higher accuracies in predicting the adversarial samples than the ANNs with the cross-entropy loss.  

\subsection{Robustness to Natural Noises}
Different from adversarial attack where the input images are affected by small, additive, classifier-tailored perturbations, natural noises add small, general, classifier-agnostic perturbations to the input images. In \cite{Dietterich2018benchmarking}, an IMAGENET-C  benchmark generated from IMAGENET\ref{Krizhevsky2012ImageNet} offers various corruption types with five severity levels for each type. 
 In this work, we apply four algorithms to generate the corrupted samples for the MINST dataset.  Assume the accuracy of the corruption type $c$ at the severity level $s(1\le s\le 5)$ for model $f$ is defined as $Acc_{s,c}^f$, and then the average accuracy is defined as the evaluation metrics:
 \begin{equation}
 Acc_c^{f}=\frac{1}{5}\sum_{s=1}^{5} Acc_{s,c}^f~.
\end{equation}
\newcommand{\tabincell}[2]{\begin{tabular}{@{}#1@{}}#2\end{tabular}} 
   \begin{table}[h]
     \begin{small}
	\centering
	\begin{tabular}{| c|  c| c| c| c| c| c| }\hline
			 \textbf{}& \tabincell{c}{Sigmoid\\(trained by BP)} &Sigmoid  &Threshold &$y=|x|$ &\tabincell{c}{Sigmoid\\(with 0-1 loss)} &\tabincell{c}{Threshold\\(with 0-1 loss)}  \\\hline
	Original & $0.96$ & $0.94$ & $0.93$  & $0.94$ & $0.84$ & $0.83$  \\\hline\hline
%	 \textbf{Corruption type}& \tabincell{c}{} & & & &\tabincell{c}{} &\tabincell{c}{}  \\\hline
	 Gaussian Noise & $0.64$ & $0.75$ & $0.75$  & $0.74$ & $0.72$ & $0.72$  \\\hline
	Impulse Noise& $0.40$ & $0.51$ & $0.51$  & $0.51$ & $0.55$ & $0.51$ \\\hline
        Glass Blur& $0.38$ & $0.44$ & $0.44$  & $0.42$ & $0.44$ & $0.43$ \\\hline
        Contrast & $0.25$ & $0.45$ & $0.39$  & $0.38$ & $0.46$ & $0.41$ \\\hline
        \textbf{Average} & $0.418$ & $0.538$ & $0.523$  & $0.513$ & $0.543$ & $0.518$ \\\hline
	\end{tabular}
	\vspace*{0.1in}
	\caption{Test robustness to natural noises for ANNs with different activation and loss functions trained by GLR. Four  types of corruption noises are adopted and the average accuracy is computed.  }
	\label{table4}
	\end{small}
\end{table}

\begin{figure}[h]
\begin{center}
\centering
\includegraphics[scale=0.2]{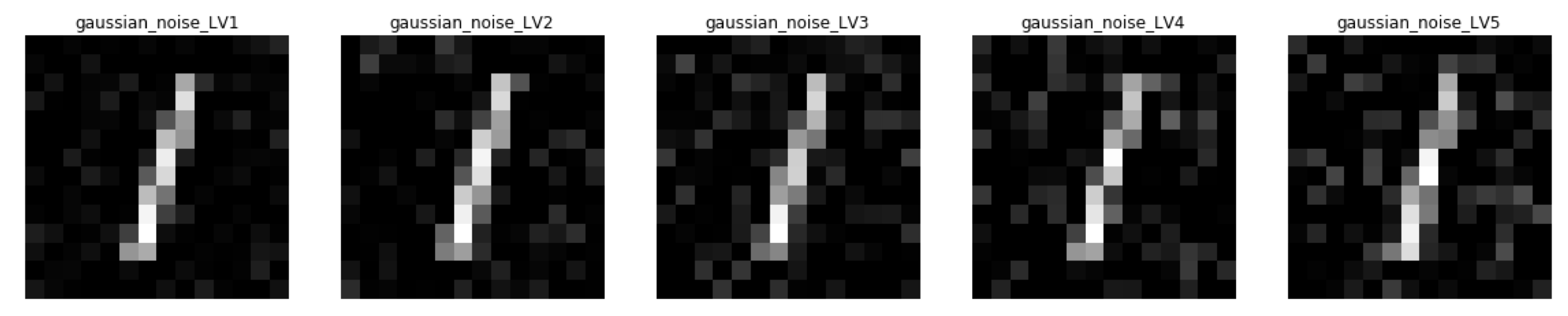}
		\caption{A sample corrupted by different levels of Gaussian noises.}
		\label{pic8} 
\end{center} 
\end{figure}

We compute the average accuracy under four types of natural noises. The  images corrupted by the Gaussian noises under five levels of severity are shown in Figure \ref{pic8}.  In Table.\ref{table4},  the ANN trained by the GLR method has a better performance than that trained by the BP method. Although the performance of the ANN with 0-1 loss is worse than that with a cross-entropy loss, an ANN with 0-1 loss achieves the best performance in predicting the images corrupted by the natural noises.

\section{Conclusions}
In this work, a GLR method is proposed for training ANNs with neuronal noises. Unlike the classic BP method, the GLR trains ANNs directly by the loss value rather than the gradient of loss and can handle ANNs with discontinuous activation and loss
functions because it does not differentiate the loss output. Therefore, the GLR method could be a powerful tool to explore some brain-like learning mechanisms which allow more freedom to better represent the surrounding environment. The robustness of all ANNs trained by the GLR method is significantly improved compared with the ANN with the Sigmoid activation function and cross-entropy loss function trained by the BP method, which indicates that the new training method is a very promising tool for enhancing the security of ANNs used in practice. 

 A future direction lies in reducing the variance of the stochastic gradient estimation for ANNs and speed up the training procedure, so that our method can be used in the deep learning ANNs with higher complexity. Adding regularization functions to the loss function for further improving robustness also deserves future research.

% In the unusual situation where you want a paper to appear in the
% references without citing it in the main text, use \nocite
\bibliographystyle{icml2019}
\bibliography{citation}

%%%%%%%%%%%%%%%%%%%%%%%%%%%%%%%%%%%%%%%%%%%%%%%%%%%%%%%%%%%%%%%%%%%%%%%%%%%%%%%
%%%%%%%%%%%%%%%%%%%%%%%%%%%%%%%%%%%%%%%%%%%%%%%%%%%%%%%%%%%%%%%%%%%%%%%%%%%%%%%
% DELETE THIS PART. DO NOT PLACE CONTENT AFTER THE REFERENCES!
%%%%%%%%%%%%%%%%%%%%%%%%%%%%%%%%%%%%%%%%%%%%%%%%%%%%%%%%%%%%%%%%%%%%%%%%%%%%%%%
%%%%%%%%%%%%%%%%%%%%%%%%%%%%%%%%%%%%%%%%%%%%%%%%%%%%%%%%%%%%%%%%%%%%%%%%%%%%%%%
%%%%%%%%%%%%%%%%%%%%%%%%%%%%%%%%%%%%%%%%%%%%%%%%%%%%%%%%%%%%%%%%%%%%%%%%%%%%%%%
%%%%%%%%%%%%%%%%%%%%%%%%%%%%%%%%%%%%%%%%%%%%%%%%%%%%%%%%%%%%%%%%%%%%%%%%%%%%%%%

\end{document}